%% file: IEEE-conference-template-062824.tex
\documentclass[a4paper,conference]{IEEEtran}
\IEEEoverridecommandlockouts

\usepackage{amsmath,graphicx,hyperref}
\usepackage{enumitem}
\usepackage{booktabs}
\usepackage{subcaption}
\usepackage{bm}
\usepackage{amsmath,amsfonts}
\usepackage{makecell}

\usepackage{soul}  

\renewcommand{\IEEEkeywordsname}{Keywords}

\usepackage{cite}
\usepackage{amsmath,amssymb,amsfonts}
\usepackage{algorithmic}
\usepackage{graphicx}
\usepackage{textcomp}
\usepackage{xcolor}

\definecolor{HL1}{HTML}{FFF2CC} 
\definecolor{HL2}{HTML}{E2F0D9} 
\definecolor{HL3}{HTML}{DDEBF7} 
\definecolor{HL4}{HTML}{FCE4D6} 
\definecolor{HL5}{HTML}{EAE1F7} 
\definecolor{HL7}{HTML}{FFD580} 

\def\BibTeX{{\rm B\kern-.05em{\sc i\kern-.025em b}\kern-.08em
    T\kern-.1667em\lower.7ex\hbox{E}\kern-.125emX}}
\begin{document}

\title{An LLM-in-the-loop RL Framework for Bioinformatics Feature Selection}


\author{\IEEEauthorblockN{Xinyuan Wang}
\IEEEauthorblockA{\textit{School of Computing and AI} \\
\textit{Arizona State University}\\
Tempe, USA \\
xwang735@asu.edu (corresponding author)}
\and
\IEEEauthorblockN{Deepti Agrawal}
\IEEEauthorblockA{\textit{School of Computing and AI} \\
\textit{Arizona State University}\\
Tempe, USA \\
dagraw18@asu.edu}
\and
\IEEEauthorblockN{Yanjie Fu}
\IEEEauthorblockA{\textit{School of Computing and AI} \\
\textit{Arizona State University}\\
Tempe, USA \\
Yanjie.Fu@asu.edu}
}

\maketitle


\input{sections/0-abstract}

\begin{IEEEkeywords}
Feature Selection, Large Language Models, Reinforcement Learning, LLM in the Loop.
\end{IEEEkeywords}

\input{sections/1-introduction}

\input{sections/2-method}

\input{sections/3-experiment}

\input{sections/4-conclusion}

\bibliographystyle{IEEEtran}
\bibliography{refs}

\end{document}

%% file: sections/0-abstract.tex
\begin{abstract}

High-dimensional bioinformatics data, characterized by a large number of features relative to the number of samples, pose major challenges such as the ``curse of dimensionality,'' leading to overfitting, high computational cost, and poor generalization.
Traditional feature selection methods often suffer from limited scalability and adaptability in such domains.
We propose an LLM-in-the-loop reinforcement learning (RL) framework for bioinformatics feature selection, where the RL agent formulates feature selection as a sequential decision-making task, while the large language model (LLM) enhances the process in two ways: (1) guiding exploration through domain-informed advice, and (2) providing hybrid rewards that integrate data-driven performance with knowledge-driven evaluation.
The LLM also produces explanations to improve interpretability for human experts without altering the RL policy update.
Experiments on diverse bioinformatics datasets show that the LLM-in-the-loop framework outperforms baselines, achieves stable performance across downstream models, and converges faster than pure RL.

\end{abstract}

%% file: sections/1-introduction.tex
\section{Introduction}
\label{sec:intro}

\begin{figure}[htbp]
    \centering
    \includegraphics[width=0.97\linewidth]{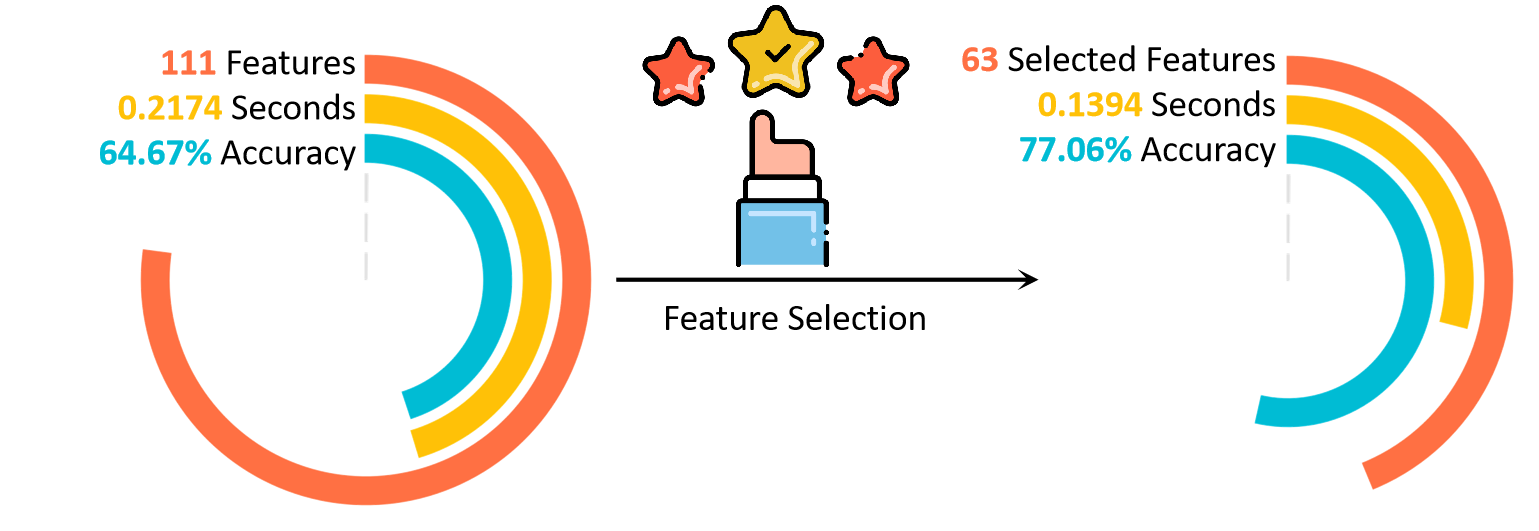}
    \caption{Feature selection brings fewer key features to improve downstream performance and reduce processing time.}
    \label{fig:motivation}
\end{figure}

With advances in data collection, high-dimensional data have become increasingly common and challenging, particularly in bioinformatics~\cite{wang2024biostatistical}. 
Such datasets often contain tens of thousands of features but only a few hundred samples in practice, giving rise to the well-known ``curse of dimensionality.'' This leads to two major challenges: (1) models are likely to overfit, as noisy correlations can outweigh true biological signals, and (2) computational costs grow rapidly, which makes model training expensive~\cite{jia2022feature}. For example, genomic or molecular datasets may require processing the full set of descriptors, which will slow down both model development and clinical translation. 
Feature selection is therefore essential to remove irrelevant or redundant features~\cite{pudjihartono2022review}. Beyond reducing complexity, selecting a biologically meaningful feature subset (e.g., key genes or biomarkers) improves predictive performance and enhances interpretability for downstream biomedical analysis and clinical research. Fig.~\ref{fig:motivation} shows how feature selection reduces downstream task processing time while boosting performance on a myocardial infarction dataset.

Traditional feature selection approaches fall into three categories: filter, wrapper, and embedded~\cite{kohavi1997wrappers}.
Filter methods (e.g., Fisher Score~\cite{hart2001pattern}, Information Gain~\cite{lewis1992feature}, Chi-Square Test~\cite{liu1995chi2}) evaluate features individually before model training. They are efficient but ignore feature interactions, which often leads to missing informative combinations.
Wrapper methods assess subsets using the performance of a predictive model~\cite{kohavi1997wrappers}. While typically more accurate, they are expensive for high-dimensional datasets, since a new model must be trained for each subset.
Embedded methods integrate selection into model training, such as Lasso regularization~\cite{tibshirani1996regression} and its extensions~\cite{zhu20031}, which enforce sparsity in parameters. These balance efficiency and accuracy, but remain model-dependent and lack generality across domains.

\begin{figure*}[htbp]
    \centering
    \includegraphics[width=0.95\linewidth]{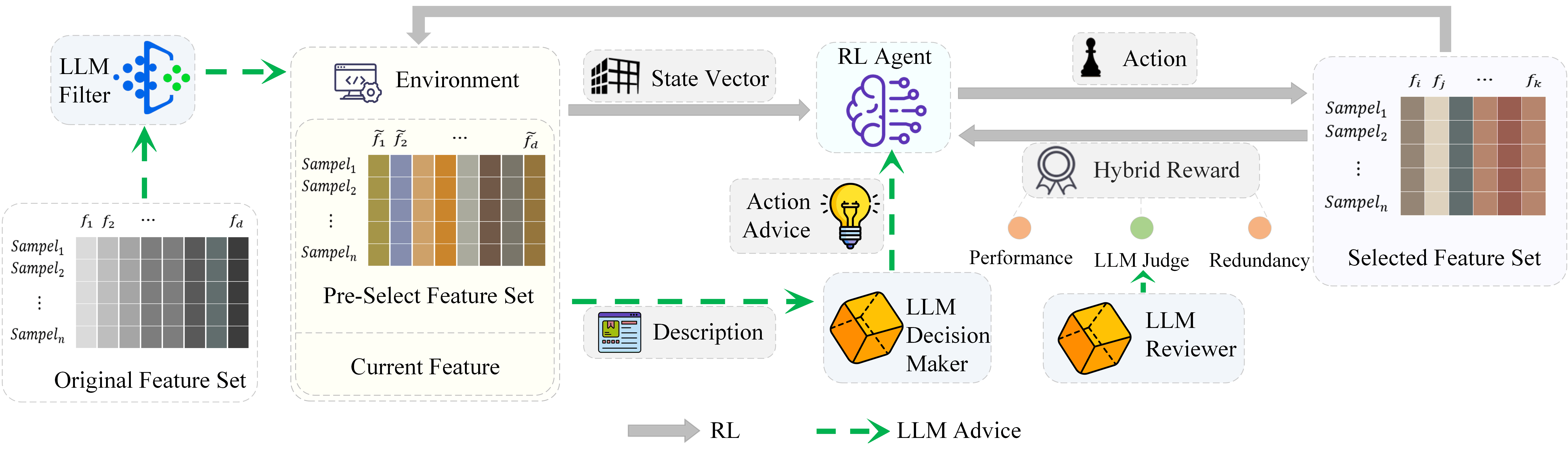}
    \caption{Overview of the LLM-in-the-loop reinforcement learning framework for feature selection.}
    \label{fig:framework}
\end{figure*}

Reinforcement Learning (RL) offers a new view on feature selection~\cite{rasoul2021feature,zhao2020simplifying,liu2019automating,wang2025knockoff}. It frames feature selection as a sequential decision process for each feature. An RL agent learns through trial and error to find the best feature subset. 
Compared to traditional methods, RL can perform a more global search and adapt to changing data environments.

RL shows great potential but faces two main challenges. 
First, exploration within large feature spaces is inefficient. The vast number of possible feature combinations makes a purely “blind” search slow and unlikely to identify the best solution. 
Second, reward signals are limited by predefined metrics. 
In RL-based feature selection, rewards are usually derived from human-defined performance metrics, which may not fully capture domain relevance or real-world utility.


To address these challenges, we propose an LLM-in-the-loop RL framework that treats LLMs as domain experts to assist the agent.
First, LLM as Decision Maker: The LLM uses its extensive domain knowledge, narrows the exploration space by suggesting promising features, and at each step advises whether the current feature should be selected.
Second, LLM as Reviewer: The LLM provides feedback after each decision, producing a hybrid reward that combines the LLM's knowledge-driven evaluation with the policy's data-driven performance. This mitigates the bias of conventional reward and offers the agent a richer and more realistic learning signal.
In addition, the LLM generates natural-language explanations of its advice and reviews, which, while not part of the RL loop, enhance interpretability for human experts.

Experiments in bioinformatics domains verify the framework. 
The results show that it achieves the best or second-best performance in most tasks, with improvements of up to 4\% in classification accuracy and a lower regression error compared to baselines. In addition, convergence analysis indicates that the LLM-in-the-loop framework achieves comparable performance more quickly, and ablation results confirm the value of LLM guidance in both decision-making and reward shaping.

%% file: sections/2-method.tex
\section{LLM-in-the-Loop RL Framework}




\subsection{Background and Task}
Feature selection is important in high-dimensional bioinformatics tasks, where datasets often contain thousands of candidate features but only limited samples. 
Using all features may introduce noise and redundancy, increase computational cost, and reduce model interpretability.

Let $\bm{D}=(\bm{X},\bm{y})$ denote a supervised dataset, where $\bm{X}\in \mathbb{R}^{n\times d}$ contains $n$ samples and $d$ features, and $\bm{y}$ denotes the labels or regression targets. 
The goal is to select a compact subset $S\subseteq\{1,\ldots,d\}$ and train a downstream model on $\bm{X}_S$, so that prediction performance is improved while redundancy and complexity are reduced.

Feature selection is combinatorial, since the number of possible subsets grows exponentially with $d$. 
RL provides a sequential formulation by deciding whether to select or skip each candidate feature. 
However, pure RL may explore inefficiently and rely only on task-driven reward signals. 
We therefore introduce LLM guidance into the RL loop, combining RL-based optimization with LLM-based knowledge about feature relevance, redundancy, and semantic consistency.

\subsection{Proposed Method}
Our framework integrates three LLM-assisted modules into the RL-based feature selection process: an LLM filter, an LLM decision maker, and an LLM reviewer. 
The LLM filter first builds a candidate priority list based on feature names, statistical summaries, and available domain descriptions. 
This step is used to prioritize exploration rather than permanently discard low-ranked features, so the RL agent can still discover useful but non-obvious candidates.

As shown in Fig.~\ref{fig:framework}, the RL agent sequentially interacts with the environment. 
At each step, the agent observes the current selected subset and the current candidate feature. 
Before making a decision, it queries the LLM decision maker, which outputs an action advice score indicating whether the candidate feature is promising. 
The RL agent combines this score with its own policy to take an action. 
After the action is executed, the environment provides downstream validation feedback, and the LLM reviewer provides an additional knowledge-driven evaluation. 
These signals are combined into a hybrid reward.

The LLM does not directly determine the final feature subset. 
Instead, it acts as a soft guide for exploration and reward shaping. 
The final policy is still learned by the RL agent using task-driven validation feedback. 
The LLM also produces short natural-language explanations for human interpretability, but these explanations are not used in policy updates.

\subsection{Reinforcement Learning Framework}
We formalize feature selection as a Markov Decision Process $(\mathcal{S}, \mathcal{A}, \mathcal{P}, R, \gamma)$:
\begin{itemize}[nosep]
    \item State $s_t \in \mathcal{S}$: the current selected subset, represented as a binary vector $\bm{x}_t \in \{0,1\}^d$, together with the current candidate feature.
    \item Action $a_t \in \mathcal{A}$: a binary decision for the current feature, i.e., $a_t \in \{\text{select}, \text{skip}\}$.
    \item Transition $\mathcal{P}(s_{t+1}|s_t,a_t)$: a deterministic update of the selected subset. If the action is \ select, the corresponding feature is added; otherwise the subset remains unchanged.
    \item Reward $R(s_t,a_t)$: a hybrid reward combining downstream validation performance and LLM-based feedback.
    \item Discount factor $\gamma$: the trade-off between short-term and long-term rewards.
\end{itemize}

The objective is to maximize the expected cumulative reward:
\begin{equation}
\pi^{*}
=
\arg\max_{\pi}
\mathbb{E}_{\pi}
\left[
\sum_{t=1}^{T}
\gamma^{t-1}
R(s_t,a_t)
\right],
\end{equation}
where $\pi$ is the policy of the RL agent and $T$ is the maximum number of decision steps. 
In our implementation, we set $T=5d$ to allow repeated exploration of uncertain features.
Already selected features are masked out during action selection, while uncertain unselected features can be revisited through the prioritized queue.

\subsubsection{LLM as Decision Maker}
The LLM decision maker provides feature-level advice before the RL agent takes an action. 
Given the current selected subset $S_t$, the candidate feature $f_t$, and task information, the LLM outputs a score $A_{\text{LLM}}(f_t,S_t)\in[0,1]$. 
A larger score indicates that the candidate feature is more likely to be useful according to domain knowledge and semantic consistency.

The LLM output follows a fixed format: a numerical score, an uncertainty flag, and a short explanation. 
Only the numerical score is used by the RL agent, while the explanation is kept for human interpretation. 
The RL agent incorporates this signal by adjusting its action probabilities:
\begin{equation}
\pi'(a_t|s_t)
\propto
\pi(a_t|s_t)
\cdot
\exp\left(\alpha g(a_t, A_{\text{LLM}}(f_t,S_t))\right),
\end{equation}
where $g(\text{select},A)=A$ and $g(\text{skip},A)=1-A$ and $\alpha$ controls the strength of LLM guidance.

LLM advice is also used in the exploration strategy. 
Besides random exploration and policy-based selection, we add an LLM-guided branch:
\begin{equation}
a_t=
\begin{cases}
\text{random action}, & \epsilon_r,\\
\arg\max_a \pi'(a|s_t), & 1-\epsilon_r-\epsilon_{\text{LLM}},\\
\arg\max_a A_{\text{LLM}}(a,s_t), & \epsilon_{\text{LLM}}.
\end{cases}
\end{equation}
This strategy preserves exploration while guiding the agent toward semantically promising features.

\subsubsection{LLM as Reviewer}
The LLM reviewer provides an intrinsic feedback signal after the agent takes an action. 
It evaluates whether the selected feature is relevant to the task, complementary to the existing subset, and not overly redundant. 
The reviewer outputs a score $u_{\text{LLM}}(s_t,a_t)\in[0,1]$, which is rescaled as
\begin{equation}
R_{\text{intrinsic}}(s_t,a_t)
=
2u_{\text{LLM}}(s_t,a_t)-1.
\end{equation}

The extrinsic reward is based on downstream validation performance and subset quality:
\begin{equation}
R_{\text{extrinsic}}(s_t,a_t)
=
\Delta Q_t
-
\beta \frac{|S_{t+1}|}{d}
-
\eta \operatorname{Red}(S_{t+1}),
\end{equation}
where $\Delta Q_t=Q(S_{t+1})-Q(S_t)$ is the validation-score improvement after action $a_t$. 
For classification, $Q(S)$ is validation accuracy; for regression, $Q(S)$ is the negative normalized $\ell_2$ error. 
The redundancy penalty $\operatorname{Red}(S)$ is computed from average pairwise feature correlation. 
When $|S|\leq 1$, we set $\operatorname{Red}(S)=0$.
The final reward is:
\begin{equation}
R(s_t,a_t)
=
\lambda R_{\text{extrinsic}}(s_t,a_t)
+
(1-\lambda)R_{\text{intrinsic}}(s_t,a_t),
\end{equation}
where $\lambda\in[0,1]$ balances validation-based feedback and LLM-based feedback. 
All reward components are normalized within each dataset to avoid scale mismatch.

\subsubsection{Policy Learning and Reliability Control}
We instantiate the RL agent with a Deep Q-Network (DQN), which outputs two action values corresponding to select and skip. 
The agent is trained using experience replay and a target network. 
For each transition $(s_t,a_t,R_t,s_{t+1})$, the temporal-difference target is
\begin{equation}
y_t
=
R_t
+
\gamma
\max_{a'}
Q_{\theta^-}(s_{t+1},a'),
\end{equation}
and the Q-network is optimized by minimizing the squared temporal-difference error.

To reduce the risk of unreliable LLM outputs, the LLM is constrained to return a score, an uncertainty flag, and a short reason. 
If the LLM response is uncertain or repeatedly conflicts with validation feedback, its influence is reduced by lowering $\alpha$ or increasing $\lambda$. 
We also cache repeated feature-level scores to reduce redundant LLM calls. 
Therefore, the LLM serves as an auxiliary knowledge source rather than an oracle, while the final feature selection policy remains driven by downstream validation performance.

%% file: sections/3-experiment.tex
\section{Experiment}

\begin{table*}[htbp]
  \begin{center}
    \caption{Comparison of the LLM-in-the-loop framework with baseline methods on bioinformatics datasets. For classification tasks, accuracy (\%) is reported (higher is better); for regression tasks, $\ell_2$ error is reported (lower is better).}
    \label{tab:overall_exp}
    \resizebox{\textwidth}{!}{\begin{tabular}{lccccccccccccc|ccc}
        \toprule
        Dataset & Task & Samples & Features & LASSO & KBest & RFE & mRMR & MCDM & MARLFS & SADRLFS & STG & FSNS & Ours & \makecell{w/o LLM \\ Action} & \makecell{w/o LLM \\ Reviewer} & \makecell{w/o LLM} \\
        \midrule
        RNA-Seq & C & 801 & 1,170 & 97.53\% & 97.53\% & 98.76\% & 98.76\% & 97.53\% & 98.76\% & 98.99\% & 97.53\% &99.31\% & 100.00\% & 98.76\% & 100.00\% & 98.96\%\\
        Tox21-NR-AR & C & 12,060 & 1,643 & 95.44\% & 96.85\% & 95.52\% & 95.61\% & 88.23\% & 96.43\% & 95.68\% & 78.94\% & 97.26\% & 97.60\% & 97.01\% & 95.35\% & 95.10\% \\
        Tox21-SR-HSE & C & 12,060 & 1,643 & 93.28\% & 93.86\% & 93.45\% & 94.94\% & 94.11\% & 95.19\% & 95.27\% & 95.10\% & 94.78\% & 95.36\% & 95.19\% & 95.11\% & 95.19\%\\
        Tox21-SR-p53 & C & 12,060 & 1,643 & 94.78\% & 89.22\% & 79.77\% & 93.11\% & 92.54\% & 95.44\% & 92.87\% & 68.91\% & 95.10\% & 95.27\% & 95.10\% & 93.69\% & 93.69\%\\
        Myocardial Infarction & C & 1,700 & 110 & 73.53\% & 75.29\% & 78.82\% & 66.47\% & 76.47\% & 77.06\% & 75.29\% & 72.35\% & 77.65\% & 79.41\% & 76.47\% & 77.06\% & 76.47\% \\
        Glycation & C & 630 & 402 & 79.37\% & 82.54\% & 77.78\% & 80.95\% & 85.71\% & 82.54\% & 87.30\% & 80.03\% & 83.70\% & 88.89\% & 85.71\% & 84.11\% & 85.71\%\\
        PimaIndian & C & 768 & 8 & 72.73\% & 71.43\% & 75.32\% & 76.62\% & 75.32\% & 75.32\% & 77.54\% & 71.43\% & 75.32\% & 80.52\% & 77.92\% & 74.03\% & 76.62\% \\
        Messidor Feature & C & 1151 & 19 & 61.21\% & 59.48\% & 67.24\% & 65.52\% & 66.38\% & 72.41\% & 69.83\% & 63.76\% & 68.81\% & 76.72\% & 71.49\% & 69.01\% & 71.55\% \\
        Yeast & C & 1,484 & 8 & 84.56\% & 85.91\% & 87.25\% & 87.25\% & 85.23\% & 87.92\% & 90.60\% & 81.47\% & 84.38\% & 87.92\% & 87.25\% & 85.23\% & 86.58\% \\
        Gallstone & C & 1,054 & 41 & 68.75\% & 71.88\% & 79.25\% & 75.00\% & 78.13\% & 75.00\% & 78.13\% & 65.63\% & 78.13\% & 81.25\% & 78.13\% & 78.13\% & 78.13\% \\
        QSAR Biodeg & C & 1,055 & 41 & 81.13\% & 85.85\% & 84.91\% & 83.02\% & 89.62\% & 86.79\% & 87.74\% & 83.96\% & 90.57\% & 91.51\% & 90.57\% & 89.62\% & 88.68\% \\
        Heart Disease & C & 270 & 13 & 64.52\% & 67.74\% & 74.19\% & 58.07\% & 64.52\% & 74.19\% & 74.19\% & 61.29\% & 77.42\% & 74.19\% & 70.97\% & 74.19\% & 67.74\% \\
        \midrule
        GSE73002 & R & 4,113 & 2,540 & 0.3556 & 0.3374 & 0.4141 & 0.3469 & 0.4069 & 0.3209 & 0.2968 & 0.4514 & 0.2849 & 0.2136 & 0.2290 & 0.2687 & 0.2352 \\
        GSE137140 & R & 3,924 & 2,371 & 0.7097 & 0.8823 & 0.6928 & 0.7679 & 0.5694 & 0.5986 & 0.5100 & 0.5848 & 0.5848 & 0.4685 & 0.5946 & 0.5925 & 0.6069\\
        GSE103322 & R & 5,902 & 755 & 1.5650 & 1.5349 & 1.5780 & 1.6018 & 1.5986 & 0.8986 & 0.9114 & 1.3417 & 0.8126 & 0.7906 & 0.8017 & 0.8465 & 0.9279\\
        GSE146026 & R & 9,609 & 497 & 1.1184 & 1.2317 & 1.1352 & 1.2031 & 1.2113 & 0.6628 & 0.7402 & 1.1436 & 0.6517 & 0.5698 & 0.5786 & 0.5811 & 0.6704\\
        GSE89567 & R & 6,341 & 797 & 1.5541 & 1.5029 & 1.4989 & 1.5565 & 1.5729 & 0.8426 & 0.8915 & 1.5323 & 0.8439 & 0.7815 & 0.8277 & 0.8438 & 1.0995\\ 
        \midrule
        Average Rank & -- & -- & -- & 7.76 & 7.06 & 5.88 & 7.06 & 6.53 & 3.94 & 3.53 & 8.18 & 3.65 & 1.41 & -- & -- & --\\ 
        \bottomrule
    \end{tabular}}
  \end{center}
\end{table*}

\subsection{Experimental Setup}

\noindent Dataset:
We evaluate on diverse datasets from bioinformatics, healthcare, and scientific domains, covering both classification and regression tasks (Table~\ref{tab:overall_exp}).
All datasets are publicly available without any identifiable personal information.

\noindent Experimental Environment:
Experiments were conducted on Ubuntu 22.04.3 with an Intel i9-13900KF CPU, Python 3.11.5, PyTorch 2.0.1, and an NVIDIA RTX 4090 GPU.

\noindent Evaluation Metrics:
For classification tasks, we report accuracy, and for regression tasks, we use the $\ell_2$ error.

\noindent Baseline Algorithms:
We compare methods from five categories: embedded (LASSO~\cite{tibshirani1996regression}, RFE~\cite{guyon2002gene}), filter (KBest~\cite{hart2001pattern}, mRMR~\cite{peng2005feature}), evolutionary (MCDM~\cite{najafi2023multiple}), neural (STG~\cite{yamada2020feature}, FSNS~\cite{gong2025neuro}), and RL (MARLFS~\cite{liu2019automating}, SADRLFS~\cite{zhao2020simplifying}).

\begin{figure}[htbp]
    \centering
    \includegraphics[width=0.85\linewidth]{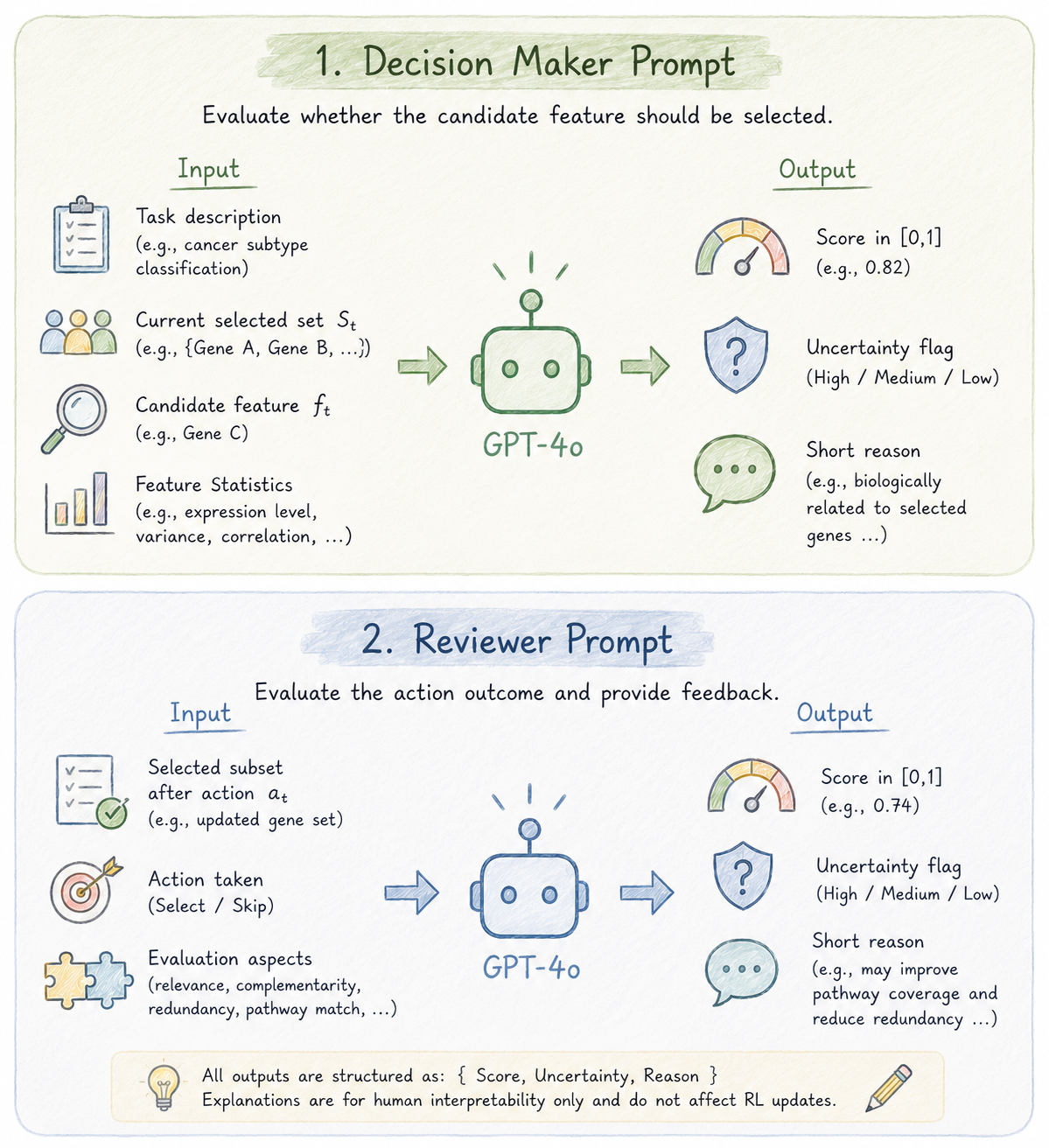}
    \caption{Prompts for the LLM decision maker and reviewer.}
    \label{fig:prompt}
\end{figure}

\noindent LLM Configuration and Prompt:
We use GPT-4o as the backbone with a fixed prompt template (Fig.~\ref{fig:prompt}). 
The LLM is used only for feature-level advice and intrinsic feedback; it does not directly determine the final selected subset. 
All LLM outputs are constrained to a numerical score, an uncertainty flag, and a short explanation.

\subsection{Performance Improvement}
Table~\ref{tab:overall_exp} gives the comparison with baseline methods on bioinformatics datasets. 
The LLM-in-the-loop framework reaches the best or second-best result on most datasets. It works better than classical filter and embedded methods (e.g., LASSO, mRMR) and also outperforms advanced approaches (e.g., STG, FSNS). 
The average rank is 1.41, showing clear improvement across both classification and regression tasks.  

The gains are clear in high-dimensional datasets such as RNA-Seq and Glycation. In these tasks, baseline methods suffer from noise and redundant features, while our framework uses LLM knowledge to focus on useful features. 
This gives up to 4\% higher classification accuracy and lower regression error. The effect is stable across datasets, showing that the framework works under different task settings.  

Insight: LLM guidance with RL improves feature selection and scales to high-dimensional bioinformatics problems.



\subsection{Ablation Study}
We test the effect of each LLM module by removing them in turn: 
(i) without the Decision Maker, 
(ii) without the Reviewer, and 
(iii) without both (pure RL). 
Table~\ref{tab:overall_exp} shows that all three cases reduce performance. The drop tends to be larger when both modules are removed.  

Without the LLM Decision Maker, the RL agent explores slowly and may miss good features. 
Without the LLM Reviewer, the reward only relies on human-defined metrics, which can be narrow or biased.
Pure RL performs worst, especially in high-dimensional data. 
Even partial use of LLM gives a gain, which means the two modules bring different but useful effects.  

Insight: The Decision Maker improves exploration by focusing on semantically relevant features, while the Reviewer refines the reward with domain-informed feedback. Together, they enable faster convergence and more accurate selection.

\subsection{Stability}
To further evaluate stability, we examine whether the proposed framework maintains consistent performance across different downstream models. 

\begin{figure}[htbp]
    \centering
    \begin{subfigure}{0.48\linewidth}
        \centering
        \includegraphics[width=\linewidth]{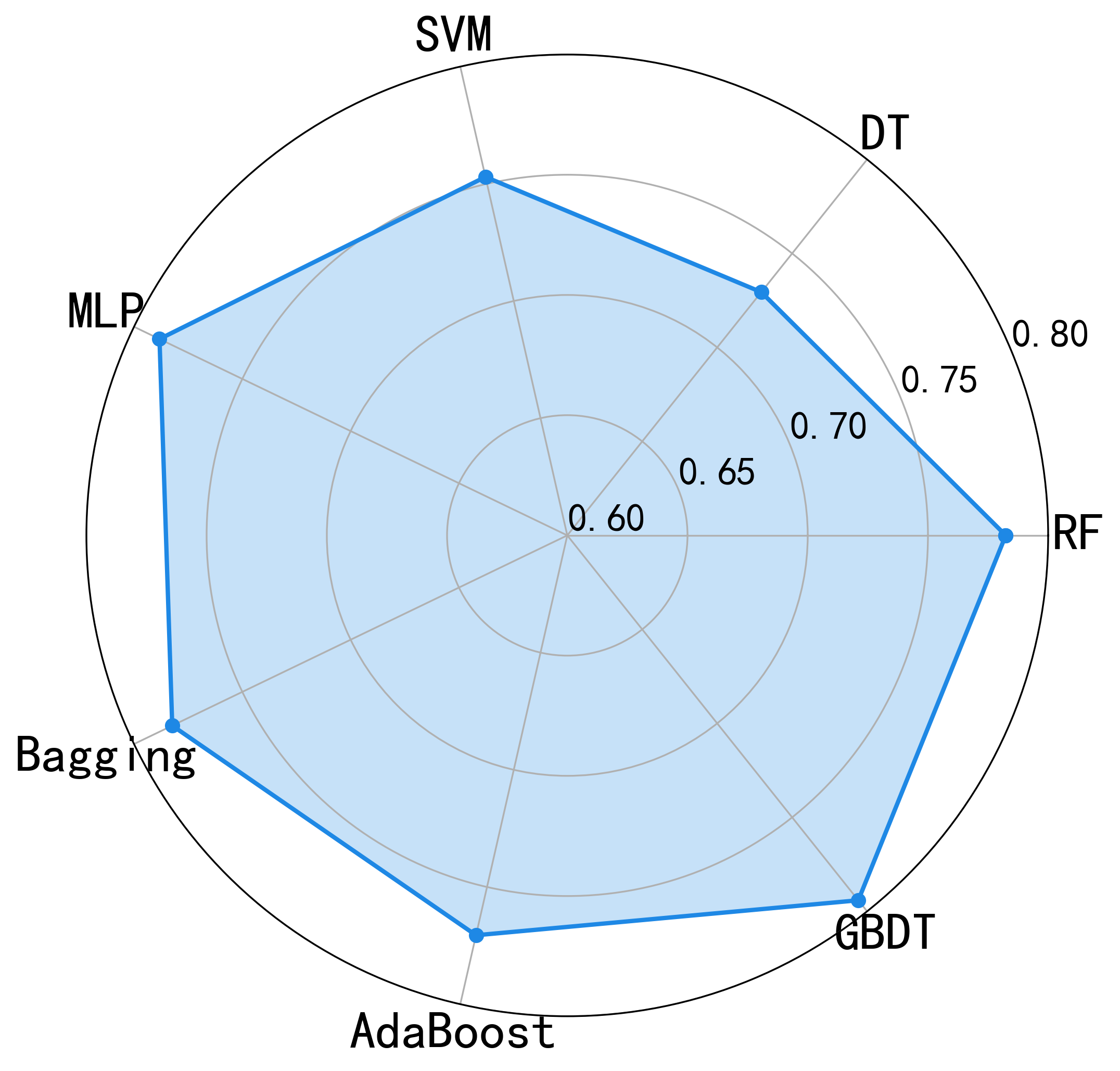}
        \caption{Myocardial Infarction}
        \label{fig:stability_cls}
    \end{subfigure}
    \hspace{0.01em} 
    \begin{subfigure}{0.48\linewidth}
        \centering
        \includegraphics[width=\linewidth]{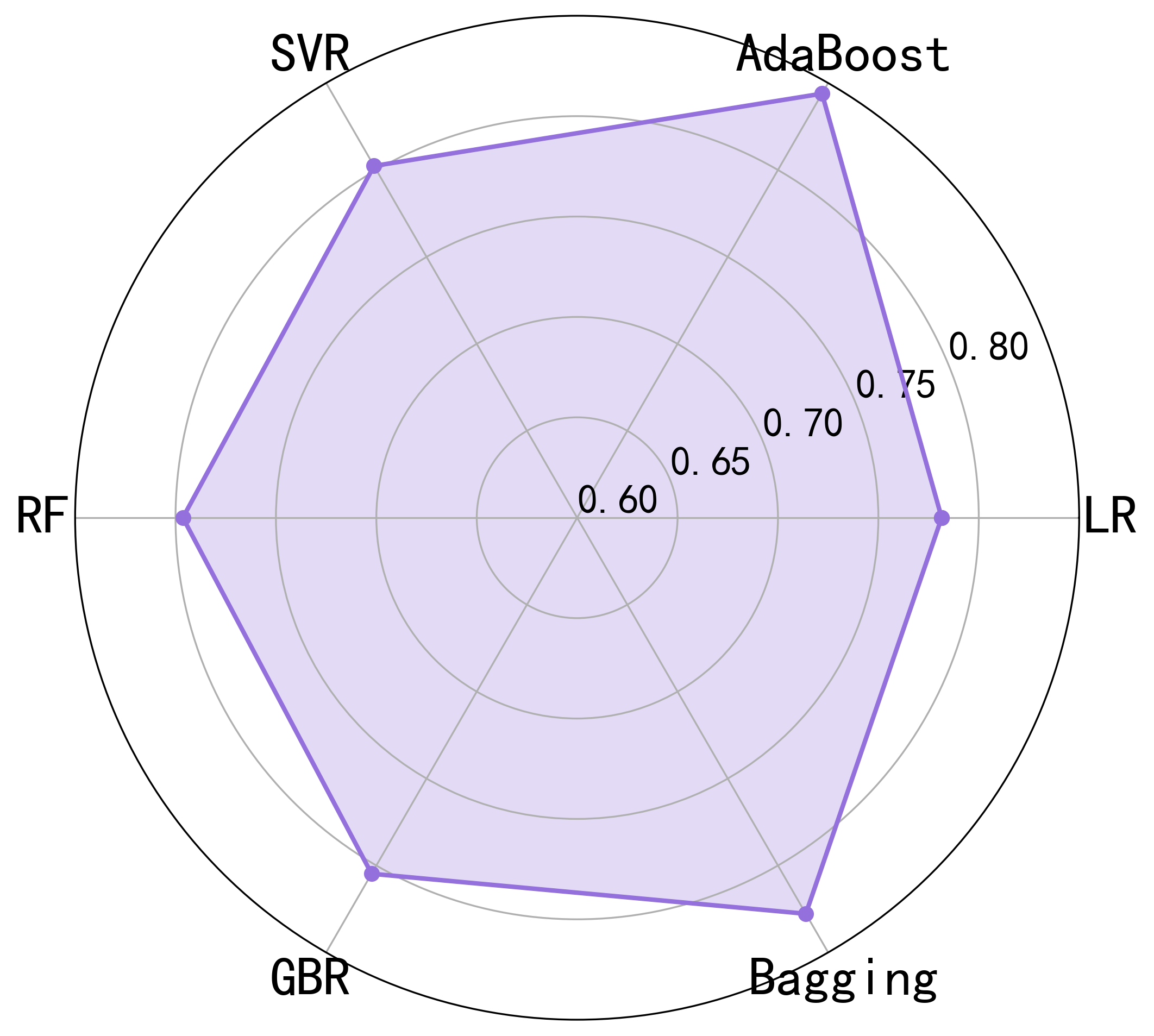}
        \caption{GSE89567 Dataset}
        \label{fig:stability_reg}
    \end{subfigure}
    \caption{Stability across downstream models.}
    \label{fig:Stability}
\end{figure}

As shown in Fig.~\ref{fig:stability_cls}, on the Myocardial Infarction dataset, the LLM-in-the-loop method achieves comparable accuracy when combined with various classifiers, including SVM, Decision Tree, and neural networks, demonstrating robustness to the choice of downstream learner. 
A similar trend is observed on regression datasets (Fig.~\ref{fig:stability_reg}), where the framework consistently delivers stable MSE across multiple regressors. 

Insight: These results indicate that the benefits of our framework with LLM guidance are not tied to a specific downstream model, but generalize well across different learning paradigms. This also confirms that the selected features form a more informative data representation, which supports better learning across tasks.

\subsection{Convergence Analysis}
To examine whether the integration of LLM guidance improves convergence efficiency compared with pure RL, we compared the run time for each round and the whole process.

\begin{figure}[htbp]
    \centering
    \begin{subfigure}{0.48\linewidth}
        \centering
        \includegraphics[width=\linewidth]{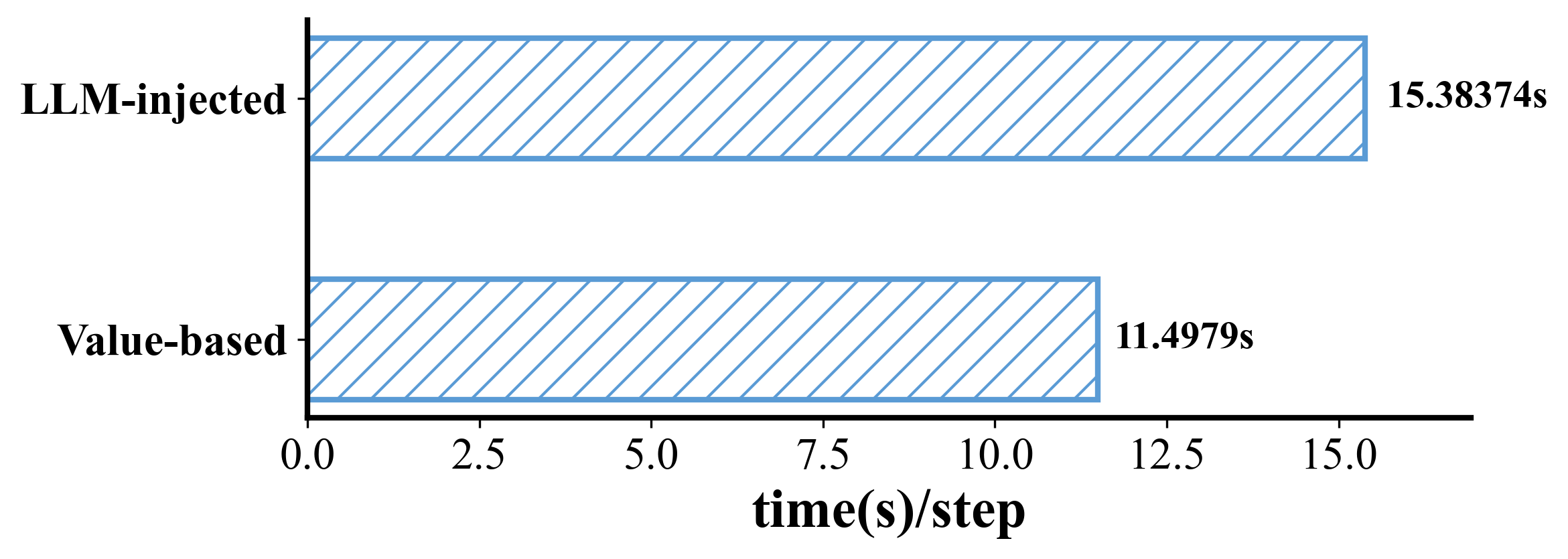}
        \caption{Run Time per Step}
        \label{fig:time_one}
    \end{subfigure}
    \hspace{0em} 
    \begin{subfigure}{0.48\linewidth}
        \centering
        \includegraphics[width=\linewidth]{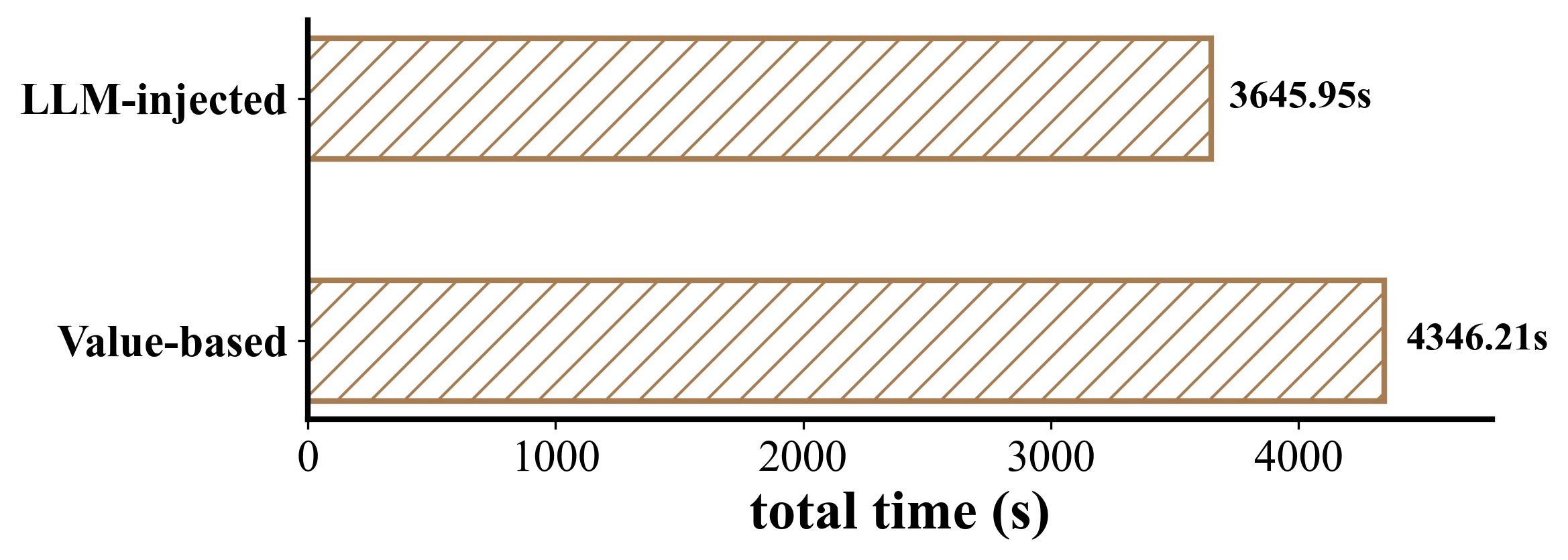}
        \caption{Total Run Time}
        \label{fig:time_all}
    \end{subfigure}
    \caption{Results on convergence efficiency.}
    \label{fig:convergence}
\end{figure}



Fig.~\ref{fig:convergence} reports the convergence efficiency results. 
As shown in Fig.~\ref{fig:time_one}, introducing LLM guidance increases the run time of each decision step because the agent needs additional LLM inference for action advice and reward feedback. 
However, Fig.~\ref{fig:time_all} shows that the proposed LLM-in-the-loop framework achieves a lower total run time than the value-based RL baseline. 
This is because the LLM-guided agent converges with fewer exploration steps, reducing repeated trials over uninformative or redundant features. 
Therefore, although LLM involvement brings extra per-step overhead, it improves the overall search efficiency of the RL process.

Insight: The convergence results reveal a useful trade-off between local computation cost and global optimization efficiency. 
Pure RL relies mainly on trial-and-error exploration, which can be inefficient in high-dimensional feature spaces. 
In contrast, our design uses the LLM decision maker to guide exploration toward semantically promising features and the LLM reviewer to provide richer reward feedback after each action. 
These two modules reduce blind exploration and help the agent reach a high-quality feature subset faster. 
Thus, the added LLM cost is compensated by more efficient policy learning and faster convergence.

\subsection{Reliability of LLM Guidance}
Since LLM outputs may be uncertain or inconsistent, we do not treat the LLM as an oracle. 
The LLM only provides auxiliary advice and intrinsic reward signals, while the final policy is still optimized using validation-based extrinsic rewards. 
To reduce hallucination risk, we use constrained prompts, fixed output formats, uncertainty flags, and consistency checks. 
If the LLM advice repeatedly conflicts with validation feedback, its influence is reduced by lowering the policy-bias weight $\alpha$ or increasing the extrinsic-reward weight $\lambda$. 
This design makes the framework robust to imperfect LLM knowledge and prevents the feature selection process from being fully controlled by the LLM.

%% file: sections/4-conclusion.tex



\section{Conclusion}
We presented an LLM-in-the-loop RL framework for feature selection, where the LLM acts as both a decision maker and a reviewer. These two roles address the key challenges of RL-based feature selection: inefficient exploration and biased reward signals. In addition, the LLM provides natural-language explanations of its advice and feedback, improving transparency and supporting human interpretability.  
Experiments on diverse bioinformatics datasets show that the LLM-in-the-loop framework outperforms baselines, achieves stable results across different downstream models, and converges faster despite the extra cost of LLM inference in high-dimensional feature selection.

%% file: refs.bib
@article{wang2024biostatistical,
  title={Biostatistical challenges in high-dimensional data analysis: Strategies and innovations},
  author={Wang, Jianjun},
  journal={Computational Molecular Biology},
  volume={14},
  number={4},
  pages={163--172},
  year={2024},
  doi={10.5376/cmb.2024.14.0019}
}

@article{jia2022feature,
  title={Feature dimensionality reduction: a review},
  author={Jia, Weikuan and Sun, Meili and Lian, Jian and Hou, Sujuan},
  journal={Complex \& Intelligent Systems},
  volume={8},
  number={3},
  pages={2663--2693},
  year={2022},
  publisher={Springer}
}

@article{pudjihartono2022review,
  title={A review of feature selection methods for machine learning-based disease risk prediction},
  author={Pudjihartono, Nicholas and Fadason, Tayaza and Kempa-Liehr, Andreas W and O'Sullivan, Justin M},
  journal={Frontiers in bioinformatics},
  volume={2},
  pages={927312},
  year={2022},
  publisher={Frontiers Media SA}
}

@article{kohavi1997wrappers,
  title={Wrappers for feature subset selection},
  author={Kohavi, Ron and John, George H},
  journal={Artificial Intelligence},
  volume={97},
  number={1-2},
  pages={273--324},
  year={1997},
  publisher={Elsevier}
}

@book{hart2001pattern,
  title={Pattern classification},
  author={Hart, Peter E and Stork, David G and Duda, RO},
  year={2001},
  publisher={Wiley Hoboken}
}

@inproceedings{lewis1992feature,
  title={Feature selection and feature extraction for text categorization},
  author={Lewis, David D},
  booktitle={Speech and Natural Language: Proceedings of a Workshop Held at Harriman, New York, February 23-26, 1992},
  year={1992}
}

@inproceedings{liu1995chi2,
  title={Chi2: Feature selection and discretization of numeric attributes},
  author={Liu, Huan and Setiono, Rudy},
  booktitle={Proceedings of 7th IEEE international conference on tools with artificial intelligence},
  pages={388--391},
  year={1995},
  organization={Ieee}
}

@article{tibshirani1996regression,
  title={Regression shrinkage and selection via the lasso},
  author={Tibshirani, Robert},
  journal={Journal of the Royal Statistical Society Series B: Statistical Methodology},
  volume={58},
  number={1},
  pages={267--288},
  year={1996},
  publisher={Oxford University Press}
}

@article{zhu20031,
  title={1-norm support vector machines},
  author={Zhu, Ji and Rosset, Saharon and Tibshirani, Robert and Hastie, Trevor},
  journal={Advances in neural information processing systems},
  volume={16},
  year={2003}
}

@inproceedings{zhao2020simplifying,
  title={Simplifying reinforced feature selection via restructured choice strategy of single agent},
  author={Zhao, Xiaosa and Liu, Kunpeng and Fan, Wei and Jiang, Lu and Zhao, Xiaowei and Yin, Minghao and Fu, Yanjie},
  booktitle={2020 IEEE International conference on data mining (ICDM)},
  pages={871--880},
  year={2020},
  organization={IEEE}
}

@inproceedings{liu2019automating,
  title={Automating feature subspace exploration via multi-agent reinforcement learning},
  author={Liu, Kunpeng and Fu, Yanjie and Wang, Pengfei and Wu, Le and Bo, Rui and Li, Xiaolin},
  booktitle={Proceedings of the 25th ACM SIGKDD International Conference on Knowledge Discovery \& Data Mining},
  pages={207--215},
  year={2019}
}

@article{rasoul2021feature,
  title={Feature selection using reinforcement learning},
  author={Rasoul, Sali and Adewole, Sodiq and Akakpo, Alphonse},
  journal={arXiv preprint arXiv:2101.09460},
  year={2021}
}

@article{gong2025neuro,
  title={Neuro-symbolic embedding for short and effective feature selection via autoregressive generation},
  author={Gong, Nanxu and Ying, Wangyang and Wang, Dongjie and Fu, Yanjie},
  journal={ACM Transactions on Intelligent Systems and Technology},
  volume={16},
  number={2},
  pages={1--21},
  year={2025},
  publisher={ACM New York, NY}
}

@inproceedings{yamada2020feature,
  title={Feature selection using stochastic gates},
  author={Yamada, Yutaro and Lindenbaum, Ofir and Negahban, Sahand and Kluger, Yuval},
  booktitle={International conference on machine learning},
  pages={10648--10659},
  year={2020},
  organization={PMLR}
}

@article{peng2005feature,
  title={Feature selection based on mutual information criteria of max-dependency, max-relevance, and min-redundancy},
  author={Peng, Hanchuan and Long, Fuhui and Ding, Chris},
  journal={IEEE Transactions on pattern analysis and machine intelligence},
  volume={27},
  number={8},
  pages={1226--1238},
  year={2005},
  publisher={IEEE}
}

@article{najafi2023multiple,
  title={Multiple-criteria decision making, feature selection, and deep learning: a golden triangle for heart disease identification},
  author={Najafi, Amirhossein and Nemati, Alireza and Ashrafzadeh, Mahdi and Zolfani, Sarfaraz Hashemkhani},
  journal={Engineering Applications of Artificial Intelligence},
  volume={125},
  pages={106662},
  year={2023},
  publisher={Elsevier}
}

@article{guyon2002gene,
  title={Gene selection for cancer classification using support vector machines},
  author={Guyon, Isabelle and Weston, Jason and Barnhill, Stephen and Vapnik, Vladimir},
  journal={Machine learning},
  volume={46},
  number={1},
  pages={389--422},
  year={2002},
  publisher={Springer}
}

@article{wang2025knockoff,
  title={Knockoff-guided feature selection via a single pre-trained reinforced agent},
  author={Wang, Xinyuan and Wang, Dongjie and Ying, Wangyang and Xie, Rui and Chen, Haifeng and Fu, Yanjie},
  journal={IEEE Transactions on Big Data},
  year={2025},
  publisher={IEEE}
}
